\documentclass{article}

\newif\ifanon
\anonfalse  % \anontrue for review, flip to \anonfalse for camera-ready

\usepackage{jtdh}
\usepackage[english]{babel}
\usepackage{url}
\usepackage[hang,flushmargin]{footmisc}
\usepackage{hyperref}
\hypersetup{colorlinks = true, linkcolor = black, urlcolor = gray, citecolor = black, anchorcolor = black}
\usepackage{booktabs}
\usepackage{amsfonts}
\usepackage{nicefrac}
\usepackage{xcolor}
\usepackage{graphicx}
\usepackage{tabularx}
\usepackage{enumitem}
\usepackage[backend=biber, style=apa, uniquelist=false]{biblatex}
\graphicspath{{img/}}

\title{Opinionated, Hesitant and Stressed:\\Three Studies of How Politicians Speak\\in Four Slavic Parliaments}

\ifanon
    \author{Anonymous Author(s)}
    \institutions{Affiliation(s) withheld for review}
\else
    \author{
    Ivan Porupski,\textsuperscript{1}\orcidlink{0009-0005-8013-8257} Nikola Ljubešić\textsuperscript{1, 2, 3}\orcidlink{0000-0001-7169-9152}
    }
    \institutions{
    \textsuperscript{1}Department of Knowledge Technologies, Jožef Stefan Institute, Ljubljana, Slovenia \rorlink{05060sz93}\par
    \textsuperscript{2}Institute of Contemporary History, Ljubljana, Slovenia \rorlink{04wr6tn02}\par
    \textsuperscript{3}Faculty of Computer and Information Science, University of Ljubljana, Slovenia \rorlink{05njb9z20}
    }
\fi

\begin{document}
\maketitle

\keywordseng{parliamentary speech, sentiment and acoustics, filled pauses, primary stress, cross-Slavic, spoken corpora, ParlaSpeech}

\section{Introduction} \label{intro}

Opinionated about what they say, hesitant about how they say it, stressed on where they put the accent. Parliamentary speech sits at an unusually rich intersection: it is planned but not scripted, public but personal, politically charged but procedurally constrained. Speakers deliver it under time pressure and adversarial attention, and what they say is inseparable from how they say it --- the pitch of a rebuttal, the hesitation before a contested figure, the stress placed on a contested word. Until recently, studying these phenomena at scale across languages has been impractical: spoken corpora have been small, mostly monolingual, and designed for automatic speech recognition rather than for linguistic or social-scientific 
\ifanon
 inquiry~\parencite{porupski2026survey-anon}.
\else
 inquiry~\parencite{porupski2026survey}.
\fi
The situation is now changing, and with it the kinds of questions one can realistically ask.

This paper pursues three such questions, each addressed to a different facet of how politicians actually speak. First, does the emotional tone of what a politician says leave a measurable trace in how they say it --- and does this trace look the same across languages? Second, what predicts when and how often speakers hesitate, and do the predictors pattern consistently across parliaments or fragment by language? Third, when a language admits multiple legitimate stress positions for the same word, is a speaker's preference for early or late stress coherent across the lexicon, or does it fragment by part of speech?

Each question is approachable today because of ParlaSpeech~\parencite{ljubevsic2025parlaspeech}, a multilingual collection of spoken parliamentary corpora covering Croatian (HR), Czech (CZ), Polish (PL), and Serbian (RS), totalling over 6{,}000 hours of speech aligned with official ParlaMint transcripts~\parencite{erjavec_parlamint_2025}. Beyond transcript alignment and rich speaker metadata, ParlaSpeech~3.0 adds Universal Dependencies annotations, utterance-level sentiment predictions, transformer-based filled pause detection, and --- for Croatian and Serbian --- word- and grapheme-level alignments with primary stress markers. These layers make the three questions above not merely askable but systematically answerable: sentiment can be cross-referenced with acoustics, filled pauses with syntactic structure, stress patterns with speaker demographics and part of speech.

We present the current state of the three investigations in Sections~\ref{sec:sentiment}--\ref{sec:primary} and then devote the remainder of the paper, Section~\ref{sec:forward}, to the research directions these studies open up.

\section{Sentiment and Acoustics} \label{sec:sentiment}

\emph{Does the emotional tone of what a politician says leave a measurable trace in how they say it?}

The intuition is old: angry speech sounds different from conciliatory speech, and listeners pick up on this long before they parse the words. Whether and how this intuition survives large-scale measurement across languages is a different question. Working with over one million utterances across all four parliaments, we examined how pitch (F0), intensity, and speech rate covary with utterance-level
\ifanon
 sentiment~\parencite{porupski2026vocal-anon}.
\else
 sentiment~\parencite{porupski2026vocal}.
\fi
Sentiment scores came from the XLM-R-ParlaSent model~\parencite{mochtak2023parlasent}; acoustic features were extracted with Praat. We tested three competing hypotheses: a linear valence pattern (negative is louder/higher/faster, positive is quieter/lower/slower), a U-shaped arousal pattern (both extremes activated, neutral flattest), or a mixture of the two. The analysis combined Wilcoxon signed-rank tests on sentiment extremes, Kendall monotonic trend analysis, and an inflection-based split to detect arousal-related curvature.

\begin{figure}[t]
    \centering
    \caption{Global feature trend averaged across pitch, intensity, and speech rate, normalised and pooled across all four languages. The minimum at sentiment value 3.5 (neutral-positive) marks the inflection point where a valence-driven decrease transitions into an arousal-related upturn.}
    \label{fig:global_trend}
    \includegraphics[width=0.6\columnwidth]{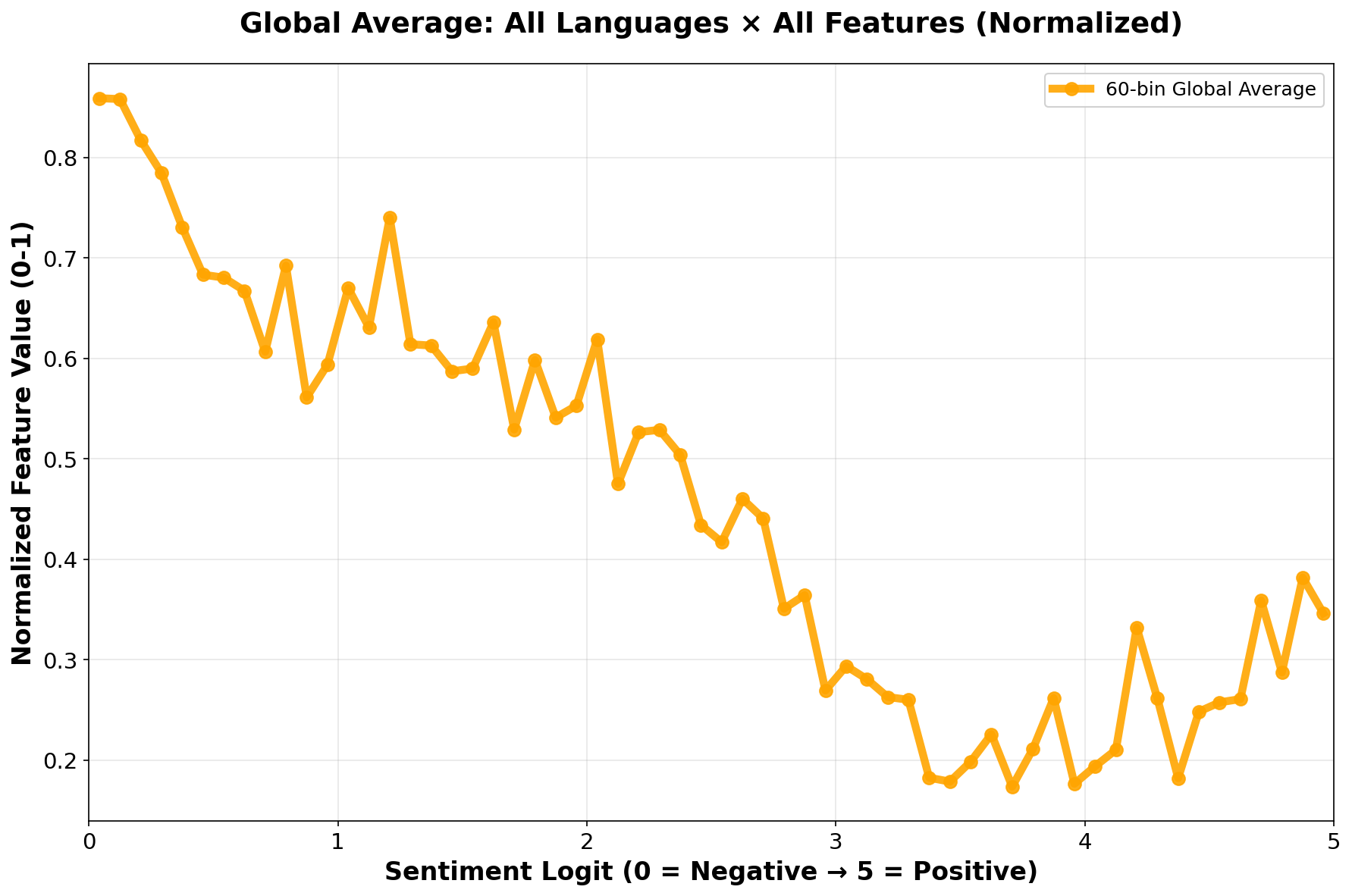}
\end{figure}

Valence dominates. In 96\% of feature--language combinations we found significant differences between sentiment extremes and monotonic trends across the full range (11/12 and 12/12 tests, respectively). As sentiment moves from negative to positive, pitch and intensity decrease systematically; speech rate shows a similar tendency, with some language-specific variation. The picture is not purely linear, however. Figure~\ref{fig:global_trend} shows a characteristic J-shape: a sustained decrease from negative through neutral sentiment, followed by a clear upturn at the most positive end of the scale. The inflection point sits at sentiment value 3.5 (the neutral-positive boundary) and is fairly consistent across features and languages. Arousal-like curvature contributes secondary modulation in 42\% of tests, most prominently for speech rate (63\% support).

Cross-linguistic differences add nuance rather than override the pattern. Croatian relies almost exclusively on linear valence tracking; Polish exhibits the strongest arousal modulation, with U-shaped patterns in pitch and intensity; Czech and Serbian sit between, with their arousal signatures concentrated in speech rate. The overall finding is that negative speech is produced with higher pitch, greater intensity, and faster rate across all four parliaments, and that the partial upturn at the positive extreme is a genuine cross-linguistic regularity rather than an artefact of any single language.

\section{Filled Pause Production} \label{sec:fp}

\emph{What predicts when and how often speakers hesitate?}

Filled pauses (``um'', ``uh'', ``eee'') sit awkwardly between disfluency and device. They can signal planning load, buy floor-holding time, or function rhetorically; they also appear to vary with age, gender, and communicative context in ways that the conversational-corpora literature has not fully reconciled. Parliamentary speech offers an unusually clean setting to probe their predictors: the domain is held constant, speaker metadata is rich, and the adversarial-yet-formal register concentrates the kind of under-pressure planning that elicits hesitation. We fitted a negative binomial GEE (Generalised Estimating Equations) model with independent working correlation to predict FP frequency per utterance (with log-duration offset), using speaker-level clusters and robust sandwich standard 
\ifanon
 errors~\parencite{porupski2026umm-anon}.
\else
 errors~\parencite{porupski2026umm}.
\fi
Filled pauses were detected via a transformer-based wav2vec2-BERT classifier~\parencite{ljubesic-etal-2025-identifying}.

\begin{figure}[t]
    \centering
    \caption{Incidence rate ratios from negative binomial GEE models with independent working correlation (baseline specification), for the pooled global model and each parliament. Error bars show 95\% confidence intervals; red markers indicate non-significant effects (CI crosses 1). Reference: female coalition speaker of average age and centrist orientation.}
    \label{fig:irr_fp_count}
    \includegraphics[width=\columnwidth]{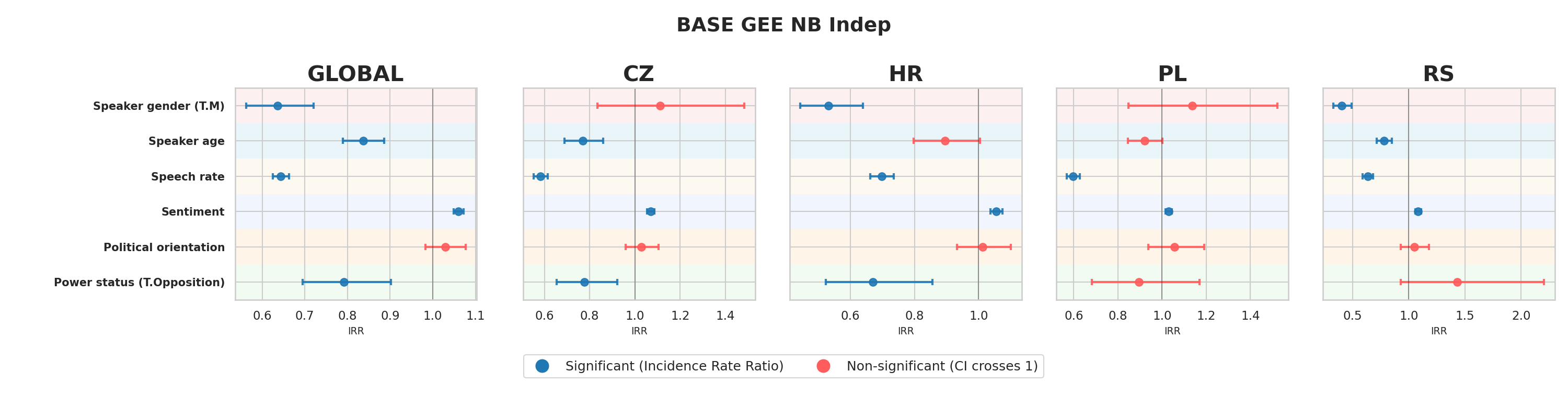}
\end{figure}

Several patterns emerge across the 1{,}001{,}787 utterances from 1{,}561 speakers in our analysis, summarised in Figure~\ref{fig:irr_fp_count}. Speech rate is the dominant predictor: faster speech yields substantially fewer FPs ($-$36\% per SD globally), consistent across all four parliaments. Sentiment shows a small but stable positive association: more positive speech corresponds to more FPs (+6\% per unit globally), stable across the full corpus. Age reduces FP frequency ($-$16\% per decade globally), clearly significant in CZ and RS (${\sim}{-}23\%$ and ${-}22\%$ per decade) with the same direction but non-significant in HR and PL. Power status shows an opposition discount globally ($-$21\%), significant in CZ ($-$22\%) and HR ($-$33\%) but non-significant in PL and borderline reversed in RS. Political orientation shows no robust global effect.

The most striking finding is cross-linguistic: gender effects reverse direction between language groups. In the South Slavic parliaments, male speakers produce substantially fewer FPs than female speakers (HR $-$47\%, RS $-$60\%); in the West Slavic parliaments (CZ, PL), there is no significant gender difference at all. This contradicts the typical ``men hesitate more'' pattern reported in conversational corpora and points to something language-, culture-, or institution-specific. A cross-linguistic design is essential for seeing it: any single-language study would either confirm or contradict the received wisdom, but only the comparison exposes the reversal as such.

%Rev3: caveat on unequal gender representation across parliaments
Women remain a numerical minority among speakers in all four legislatures; the clustered, robust-SE specification reflects the resulting sample-size asymmetry in wider confidence intervals for the less-represented group, but the underlying imbalance is worth bearing in mind when interpreting the size of the gender effect rather than only its significance.

%Rev2: interpretive discussion of the gender reversal finding
Several non-exclusive explanations could underlie this split: South Slavic parliamentary cultures may carry stronger gendered expectations around confident, disfluency-free speech, suppressing or penalising hesitation more for one gender than the other; floor-time allocation and seniority structures differ across these four legislatures in ways that could shape who speaks with more rehearsed fluency; and broader public-speaking norms outside parliament may compound the effect if they diverge between South and West Slavic public life. We cannot adjudicate between these accounts with the current data, and see this as a natural target for the political-rhetoric programme sketched in Section~\ref{fw:rhetoric}.

\section{Primary Stress in Croatian} \label{sec:primary}

\emph{Is a speaker's preference for early vs.\ late stress consistent across the lexicon, or does it fragment by part of speech?}

\begin{figure}[h]
    \centering
    \caption{MDS mapping into 2D of pairwise correlation coefficients between speaker-specific preferences for early vs.\ late stress across parts of speech.}
    \label{fig:mds}
    \includegraphics[width=0.55\columnwidth]{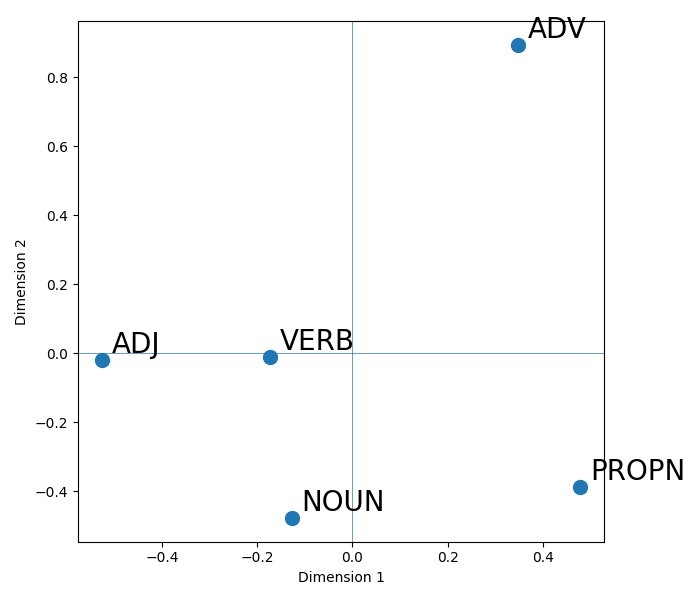}
\end{figure}

Croatian has a well-documented accentual variation: the same multisyllabic word often admits two legitimate stress positions (e.g.\ \emph{n\'agla\v{s}avam} vs.\ \emph{nagla\v{s}\'avam}), and the distribution of early and late patterns varies across the lexicon. Descriptive work by~\textcite{ivic1994} treats Neo-\v{S}tokavian accentuation as structurally constrained but allows for paradigm-internal alternations, while~\textcite{kapovic2015} foregrounds diachronic stratification and analogical levelling. More recent experimental work~\parencite{pletikos2019croatian} shows that variation is gradient, shaped by frequency, morphology, and register. What none of this prior work has been able to address is whether speaker-level preferences cohere across the lexicon: when a given speaker favours early stress in verbs, do they also favour it in nouns, adjectives, adverbs, and proper names?

Every multisyllabic word in the Croatian portion of ParlaSpeech was automatically annotated for primary stress using a fine-tuned speech transformer achieving 99\% word-level accuracy~\parencite{ljubesic2025primary_stress}. %Rev1: clarifies Croatian-only scope (Serbian alignments exist too, see intro)
We focus on Croatian here because its stress-annotation pipeline was the first to be validated at scale; the same alignments exist for Serbian (Section~\ref{intro}), and extending the present analysis to that language is discussed in Section~\ref{fw:stress}. From 11.3 million multisyllabic tokens (496 speakers), we retained words exhibiting stress variation --- those where the secondary stress position accounted for at least 10\% of occurrences --- leaving 935k tokens from 491 speakers. Each speaker was represented by their proportion of early stress usage per part of speech.

Pairwise Pearson correlations across parts of speech reveal a tight central cluster of verbs, adjectives (r = 0.88***), and nouns (r = 0.84***). Proper nouns are only partially aligned with this cluster (r = 0.63--0.70***), and adverbs lie largely outside it (r = 0.23--0.27***). Adverbs and proper nouns are mutually near-independent (r = 0.04, n.s.). The MDS projection in Figure~\ref{fig:mds} summarises the picture: speaker-level stress preferences are coherent within an open-class core (verbs--adjectives--common nouns) but decouple for adverbs and proper nouns. This suggests that ``a speaker's stress system'' is not a single setting but a structured bundle, and that adverbs in particular behave more like lexical idiosyncrasies than like instantiations of a general preference.

\section{Future Work} \label{sec:forward}

The three studies above each open onto larger programmes, which we sketch along five axes: corpus expansion, corpus phonetics, disfluencies, primary stress variation, and political rhetoric.

\subsection{Corpus Expansion and Additional Annotation Layers} \label{fw:expansion}

Slovenian is the immediate next language addition: the pipeline extends naturally and the linguistic proximity to the existing languages makes it methodologically tractable. Bosnian, additional Serbian varieties, Bulgarian, and Ukrainian are strong further candidates, each bringing distinct typological or sociolinguistic leverage. On the annotation side, silent pauses are the obvious complement to filled pauses --- inter-word silences can be extracted directly from the existing word-level alignments at essentially no additional cost --- and the filled pause inventory should be further subclassified by phonetic type (schwa-like, eee-like, and so on). Topic labels from the ParlaCAP project would add a policy-domain axis, enabling everything from within-speaker register comparisons to topic-controlled acoustic analysis.

\subsection{Corpus Phonetics} \label{fw:phonetics}

The phonetic description of the four languages represented in ParlaSpeech rests heavily on classical lab-based studies with small speaker samples and read speech. Corpus linguistics transformed syntax and lexicography by making large naturalistic samples the primary empirical basis; a corpus-phonetics turn is now possible for these languages, and the precise grapheme- and phoneme-level alignments available for Croatian and Serbian make it immediately tractable. Vowel spaces can be re-measured at population scale by extracting F1/F2 from every vowel token and comparing the empirical space to the textbook five-vowel system, with conditioning on gender, age, and speech rate. The degree of vowel reduction in connected speech, often claimed to be limited in South Slavic relative to (say) Russian, becomes a measurable rather than an asserted property. Phoneme duration inventories are recoverable not just as means but as full distributions, allowing the phonological short/long vowel contrast to be assessed for how cleanly it separates in spontaneous speech.

More targeted comparisons follow. Voice onset time in plosives, classically described as prevoiced for voiced stops and short-lag for voiceless in Croatian and Serbian, can be verified at scale and checked for stability across speech rates and speaker demographics. Consonant cluster simplification in fast speech can be systematically measured, with the set of simplifying elements compared to predictions from the phonological literature. The Croatian--Serbian comparison is especially attractive: two closely related varieties processed through the same annotation pipeline mean that any measured differences are genuinely linguistic rather than methodological. And across all of these measurements, conditioning on speech rate yields a hyper-/hypo-articulation profile --- a characterisation of what gives first as speakers speed up: vowel quality, consonant duration, the voicing contrast, or some interaction.

\subsection{Disfluencies} \label{fw:disfluency}

Filled pauses are only one dimension of the hesitation story, and the frequency analysis reported in Section~\ref{sec:fp} is only one angle on filled pauses themselves. A natural question the existing annotation layers put within reach is whether filled pauses cluster at particular positions in the syntactic tree --- for instance between rather than within phrases --- and whether any such pattern is consistent across parliaments. An information-theoretic complement asks whether surprisal rises following a filled pause: related work suggests it does, but the generality of the effect remains open. A further dimension, foreshadowed in the frequency analysis, is FP duration: frequency and duration are separable aspects of disfluency whose predictor profiles need not coincide --- a cleavage we are pursuing in follow-up work. Extending the inventory further to repetitions and repairs alongside filled and silent pauses would complete the spoken-fluency picture.

Once silent pauses are extracted, the full pausing profile becomes tractable: speaker-level trends in the ratio of filled to silent pauses, co-occurrence statistics, and whether the predictors identified for FP frequency transfer to silent pauses or pattern differently. The timing information in the alignment layer also opens questions beyond disfluency proper --- redefining multi-word expressions as sequences whose inter-word pauses are systematically shorter than expected given length and frequency, and reassessing contested syntactic boundaries via pause distribution across candidate parses. In a complementary direction, ongoing collaboration with Croatian psycholinguists examines word production duration as a function of concreteness and imageability ratings, with scope for extension to valence and arousal norms and to the other languages as such norms become available.

\subsection{Primary Stress Variation} \label{fw:stress}

%Rev1: names Serbian as the immediate next step given existing alignments
The most immediate extension is to Serbian, for which the same word- and grapheme-level stress alignments already exist (Section~\ref{intro}); replicating the coherence analysis there would show whether the verbs--adjectives--nouns cluster and the adverb/proper-noun periphery reported for Croatian reflect a general Neo-\v{S}tokavian pattern or something specific to Croatian speakers.

The analysis in Section~\ref{sec:primary} treats parts of speech at the broadest grain; subcategory analysis is the obvious next step, asking whether the verbs--adjectives--nouns coherence holds uniformly across aspect, tense, voice, or declension class. A more ambitious direction is to classify speakers into Croatian's two main systems --- pitch-accent and stress --- and use the classification to probe several hypotheses: that stress-system speakers deviate from their own system more often than pitch-system speakers deviate from theirs, that deviations are more lexically specific than context-driven, and that temporal trends within sessions reveal accommodation to previous speakers. Overcorrection is a related target: stress-system speakers overgeneralising early stress onto words where early stress is non-standard, and --- symmetrically --- pitch-system speakers applying late stress where stress-system speakers do not.

\subsection{Political Rhetoric} \label{fw:rhetoric}

Parliamentary speech is a political act, and the acoustic and disfluency signals examined above are also potentially rhetorical resources. Do speakers shift pitch, rate, and intensity when moving between policy domains --- defence versus healthcare, for instance --- independently of sentiment? Do the same politicians sound measurably different in government versus opposition, when the same individual can be tracked across sessions on either side of a transition? Does acoustic accommodation to the previous speaker show up within debate sessions, generalising the accommodation hypothesis from primary stress to the full acoustic profile? And can filled pauses be decomposed into planning-related and strategic-rhetorical subtypes using syntactic tree position and the sentiment trajectory of their surrounding context? Each of these is a research question in its own right, and together they point toward a treatment of parliamentary speech as a site for studying the rhetorical exploitation of paralinguistic resources.

%Rev3: sociolinguistic/historical expertise + procedural heterogeneity (scripted vs free speech)
Pursuing these questions well will benefit from expertise beyond corpus phonetics. Sociolinguists and historians are better placed than we are to characterise how the four parliaments' distinct historical trajectories and institutional cultures shape norms of public speech, and procedural heterogeneity we have not yet modelled --- most obviously the balance between scripted, pre-written delivery and extemporaneous response, which plausibly varies by parliament, party, and debate type --- is a natural candidate for future conditioning variables once reliably annotated.

\section{Conclusion} \label{sec:conclusion}

The three studies presented here use the same corpus and broadly the same methodological toolkit, but they address very different kinds of question: a cross-modal one about how sentiment is acoustically realised, a behavioural one about who hesitates and when, and a structural one about whether a speaker's phonological preferences hold together across the lexicon. What unites the findings is that parliamentary speech, despite its formal register and genre constraints, is a rich site for observing how language, emotion, and politics interact, and that the patterns one sees are neither uniform across languages nor random with respect to them. Negative sentiment recruits a consistent acoustic profile --- higher, louder, faster --- across all four parliaments, with a cross-linguistically stable upturn at the most positive end of the scale. The predictors of hesitation rate pattern broadly similarly across languages except along the gender axis, where South and West Slavic parliaments diverge sharply. Croatian speakers' stress preferences cohere within a verbal core but fragment at the periphery, complicating any account that treats ``speaker system'' as unitary.

Taken together, these results argue for parliamentary speech as a productive domain for comparative research on the intersection of language, affect, and political behaviour, and they argue more specifically for cross-linguistic designs in which language-level variation is part of the signal rather than something to be controlled away. The forward programme sketched in Section~\ref{sec:forward} suggests that the findings reported here are preliminary in a useful sense: each answers its question while opening several new ones that the same corpus, suitably extended, can be used to pursue. Building further large, open, comparable, richly annotated spoken datasets --- across additional languages, annotation layers, and parliamentary contexts --- is the natural way to turn these openings into a sustained research programme.

\acknowledgmenteng{
This work was supported in part by the project ``Large Language Models for Digital Humanities'' (Grant GC-0002), the project ``EPIC-SI - Early Parent-Child Communication in Slovenian: Corpus-based Insights'' (J6-70222), the research programme ``Language Resources and Technologies for Slovene'' (Grant P6-0411), and the Research Infrastructure DARIAH-SI (I0-E007), all funded by the ARIS Slovenian Research and Innovation Agency.
}

\urlstyle{same}
\printbibliography

\end{document}